\documentclass{article} 
\usepackage{iclr2021_conference,times}

\usepackage[colorlinks=true, linkcolor=black, citecolor=darkgray]{hyperref}

\usepackage{amsmath,amsfonts,bm}

\def\eqref#1{equation~\ref{#1}}

\def\1{\bm{1}}

\def\vzero{{\bm{0}}}

\def\vg{{\bm{g}}}

\def\vx{{\bm{x}}}

\def\vz{{\bm{z}}}

\def\mJ{{\bm{J}}}

\DeclareMathAlphabet{\mathsfit}{\encodingdefault}{\sfdefault}{m}{sl}
\SetMathAlphabet{\mathsfit}{bold}{\encodingdefault}{\sfdefault}{bx}{n}

\def\gA{{\mathcal{A}}}

\def\gF{{\mathcal{F}}}

\def\gM{{\mathcal{M}}}

\def\gO{{\mathcal{O}}}

\def\gX{{\mathcal{X}}}

\def\gZ{{\mathcal{Z}}}

\usepackage{graphicx}
\usepackage[separate-uncertainty=true]{siunitx}
\usepackage{booktabs}       
\usepackage{amsfonts,amsmath,amssymb,amsthm}       
\usepackage{nicefrac}       
\usepackage{mathtools}
\usepackage{listings}
\usepackage{dsfont}
\usepackage{bm}
\usepackage{rotating}
\usepackage{svg}
\usepackage{hyperref}
\usepackage{xcolor}
\usepackage{url}            
\usepackage[ruled]{algorithm2e}
\usepackage{tikz}
\usetikzlibrary{calc}

\makeatletter
\def\mathcolor#1#{\@mathcolor{#1}}
\def\@mathcolor#1#2#3{%
  \protect\leavevmode
  \begingroup
  \color#1{#2}#3%
  \endgroup
}
\makeatother

\newcommand{\wrt}{w.r.t.~}

\newcommand{\beq}[1]{\begin{equation} \eqlab{#1}}
\newcommand{\eeq}{\end{equation}}
\newcommand{\bal}{\begin{align}}
\newcommand{\eal}{\end{align}}
\newcommand{\balnn}{\begin{align*}}
\newcommand{\ealnn}{\end{align*}}

\newcommand{\bsub}{\begin{subequations}}
\newcommand{\esub}{\end{subequations}}

\newcommand{\eqlab}[1]{\label{eq:#1}}
\renewcommand{\eqref}[1]{Eq.~(\ref{eq:#1})}

\newcommand{\grad}{\boldsymbol{\nabla}}

\newcommand{\pp}{\partial}

\renewcommand{\vec}[1]{\bm{#1}}
\newcommand{\trans}[1]{#1^\mr{T}}

\newcommand{\real}{\mathbb{R}}

\newcommand{\br}[1]{\left\lbrack #1 \right\rbrack}
\newcommand{\paren}[1]{\left(#1\right)}
\newcommand{\tub}[1]{\left\{#1\right\}}
\newcommand{\mr}[1]{\mathrm{#1}}

\newcommand{\norm}[1]{\left\lVert#1\right\rVert}

\newcommand{\xxx}{\vec{x}}

\newcommand{\XXX}{\vec{X}}
\newcommand{\XXXt}{\trans{\XXX}}

\newcommand{\mean}[1]{\mathbb{E}\br{#1}}

\newcommand{\meanp}[2]{\mathbb{E}_{#1}\br{#2}}

\usepackage{url}
\usepackage{enumitem}
\usepackage[capitalise]{cleveref}
\renewcommand{\theenumi}{\arabic{enumi}}
\renewcommand{\theenumii}{\arabic{enumii}}

\makeatletter
\renewcommand{\p@enumii}{\theenumi.}
\renewcommand{\p@enumiii}{\theenumi.\theenumii.}
\makeatother
\usepackage[colorinlistoftodos,prependcaption, textsize=tiny,
 textwidth=2cm]{todonotes}
\usepackage{subcaption}
\usepackage{caption}
\usepackage{wrapfig}
\newtheorem{corollary}{Corollary}

\newtheorem{theorem}{Theorem}

\title{Assisting the Adversary \\ to Improve GAN Training}

\author{Andreas Munk, William Harvey \& Frank Wood  \\
Department of Computer Science, \\
University of British Columbia, \\
Vancouver, Canada \\
\texttt{\{amunk,wsgh,fwood\}@cs.ubc.ca} \\
}

\def\wgan{WGAN-GP}
\def\gorig{\vg_{\mr{orig}}}
\def\gadvas{\vg_{\mr{\advas{}}}} \def\gtotal{\vg_{\mr{total}}}

\def\advas{AdvAs}  

\newcommand{\madvas}{r(\theta,\phi)}

\newcommand{\ladv}{L_{\mr{adv}}}
\newcommand{\lgen}{L_{\mr{gen}}}
\newcommand{\lgenadvas}{L^{\mr{\advas{}}}_{\mr{gen}}}

\newcommand{\ptrue}{p_{\mr{true}}}

\newcommand{\nadv}{n_{\mr{adv}}}

\iclrfinalpre 
\begin{document}

\maketitle

\begin{abstract}
  Some of the most popular methods for improving the stability and performance
  of GANs involve constraining or regularizing the discriminator. In this paper
  we consider a largely overlooked regularization technique which we refer to as
  the Adversary's Assistant (\advas{}). We motivate this using a different
  perspective to that of prior work. Specifically, we consider a common mismatch
  between theoretical analysis and practice: analysis often assumes that the
  discriminator reaches its optimum on each iteration. In practice, this is
  essentially never true, often leading to poor gradient estimates for the
  generator. To address this, \advas{} is a theoretically motivated penalty
  imposed on the generator based on the norm of the gradients used to train the
  discriminator. This encourages the generator to move towards points where the
  discriminator is optimal. We demonstrate the effect of applying \advas{} to
  several GAN objectives, datasets and network architectures. The results
  indicate a reduction in the mismatch between theory and practice and that
  \advas{} can lead to improvement of GAN training, as measured by FID scores.
\end{abstract}

\section{Introduction}

The generative adversarial network (GAN) framework
~\citep{goodfellow2014generative} trains a neural network known as a
\textit{generator} which maps from a random vector to an output such as an
image. Key to training is another neural network, the \textit{adversary}
(sometimes called a discriminator or critic), which is trained to distinguish
between ``true'' and generated data. This is done by \textit{maximizing} one of
the many different objectives proposed in the literature; see for
instance~\citet{goodfellow2014generative,arjovsky2017wasserstein,nowozin2016fgan}.
The generator directly competes against the adversary: it is trained to
\textit{minimize} the same objective, which it does by making the generated data
more similar to the true data. GANs are efficient to sample from, requiring a
single pass through a deep network, and highly flexible, as they do not require
an explicit likelihood. They are especially suited to producing photo-realistic
images~\citep{zhou2019hype} compared to competing methods like normalizing
flows, which impose strict requirements on the neural network
architecture~\citep{kobyzev2020normalizing,rezende2015variational} and
VAEs~\citep{kingma2014autoencoding,razavi2019generating,vahdat2020nvae}.
Counterbalancing their appealing properties, GANs can have unstable training
dynamics~\citep{kurach2019largescale,goodfellow2017nips,kodali2017convergence,mescheder2018which}.

Substantial research effort has been directed towards improving the training of
GANs. These endeavors can generally be divided into two camps, albeit with
significant overlap. The first develops better learning objectives for the
generator/adversary to minimize/maximize. These are designed to have properties
which improve training~\citep{arjovsky2017wasserstein,li2017mmd,nowozin2016fgan}.
The other camp develops techniques to regularize the adversary and improve its
training
dynamics~\citep{kodali2017convergence,roth2017stabilizing,miyato2018spectral}.
The adversary can then provide a better learning signal for the generator.
Despite these contributions, stabilizing the training of GANs remains unsolved
and continues to be an active research area.

An overlooked approach is to train the generator in a way that accounts for the
adversary not being trained to convergence. One such approach was introduced by
\citet{mescheder2017numerics} and later built on by
\citet{nagarajan2017gradient}. The proposed method is a regularization term
based on the norm of the gradients used to train the adversary. This is
motivated as a means to improve the convergence properties of the minimax game.
The purpose of this paper is to provide a new perspective as to why this
regularizer is appropriate. Our perspective differs in that we view it as
promoting updates that lead to a solution that satisfies a sufficient condition
for the adversary to be optimal. To be precise, it encourages the generator to
move towards points where the adversary's current parameters are optimal.
Informally, this regularizer ``assists'' the adversary, and for this reason we
refer to this regularization method as the \textit{Adversary's Assistant}
(\advas{}).

We additionally propose a version of \advas{} which is hyperparameter-free.
Furthermore, we release a library which makes it simple to integrate into
existing code. We demonstrate its application to a standard architecture with
the WGAN-GP objective~\citep{arjovsky2017wasserstein,gulrajani2017improved}; the
state-of-the-art StyleGAN2 architecture and objective introduced by
\citet{karras2020analyzing}; and the AutoGAN architecture and objective
introduced by \citet{gong2019autogan}. We test these on the
MNIST~\citep{lecun1998gradientbased}, CelebA~\citep{liu2015deep},
CIFAR10~\citep{krizhevsky2009learning} datasets respectively. We show that
\advas{} improves training on all datasets, as measured by the Fr\'echet
Inception Distance (FID)~\citep{heusel2017gans}, and the inception
score~\citep{salimans2016improved} where applicable.

\section{Background}
A generator is a neural network $g:\gZ\rightarrow\gX\subseteq\real^{d_{x}}$
which maps from a random vector $\vz\in\gZ$ to an output $\vx \in \gX$ (e.g., an
image). Due to the distribution over $\vz$, the function $g$ induces a
distribution over its output $\vx=g(\vz)$.
If $g$ is invertible and differentiable, the probability density function (PDF)
over $\vx$ from this ``change of variables'' could be computed. This is not
necessary for training GANs, meaning that no such restrictions need to be placed
on the neural network $g$.
We denote this distribution $p_{\theta}(\vx)$ where
$\theta\in\Theta\subseteq\mathbb{R}^{d_g}$ denotes the generator's parameters.
The GAN is trained on a dataset $\vx_1,\ldots,\vx_N$, where each $\vx_i$ is in
$\gX$. We assume that this is sampled i.i.d. from a data-generating distribution
$\ptrue{}$. Then the aim of training is to learn $\theta$ so that $p_{\theta}$
is as close as possible to $\ptrue{}$. Section~\ref{sec:gan-objectives} will
make precise what is meant by ``close.''


The adversary $a_\phi:\gX\rightarrow \gA$ has parameters
$\phi\in\Phi\subseteq\real^{d_a}$ which are typically trained alternately with
the generator. It receives as input either the data or the generator's outputs.
The set that it maps to, $\gA$, is dependent on the GAN type.
For example, \citet{goodfellow2014generative} define an adversary which maps
from $x\in\gX$ to the probability that $x$ is a ``real'' data point from the
dataset, as opposed to a ``fake'' from the generator. They therefore choose
$\gA$ to be $[0, 1]$ and train the adversary by maximizing the associated
log-likelihood objective,
\begin{equation}
  h(p_{\theta},a_\phi) = \mathbb{E}_{x\sim \ptrue} \left[ \log a_\phi(x) \right] + \mathbb{E}_{x\sim p_{\theta}} \left[ \log (1 - a_\phi(x)) \right].
  \label{eq:original-h}
\end{equation}
Using the intuition that the generator should generate samples that seem real
and therefore ``fool'' the adversary, the generator is trained to minimize
$h(p_{\theta},a_\phi)$. Since we find $\theta$ to minimize this objective while
fitting $\phi$ to maximize it, training a GAN is equivalent to solving the
minimax game,
\begin{equation}
  \label{eq:general-form}
 \min_{\theta}\max_{\phi}h(p_{\theta},a_\phi).
\end{equation}
\cref{eq:original-h} gives the original form for $h(p_{\theta},a_\phi)$ used by
\citet{goodfellow2014generative} but this form varies between different GANs, as
we will discuss in \cref{sec:gan-objectives}. The minimization and maximization
in \cref{eq:general-form} are performed with gradient descent in practice. To be
precise, we define $L_{\text{gen}}(\theta, \phi) = h(p_{\theta},a_\phi)$ and
$L_{\text{adv}}(\theta, \phi) = -h(p_{\theta},a_\phi)$. These are treated as
losses for the generator and adversary respectively, and both are minimized. In
other words, we turn the maximization of $h(p_{\theta},a_\phi)$ \wrt $\phi$ into
a minimization of $L_{\text{adv}}(\theta, \phi)$. Then on each iteration,
$\theta$ and $\phi$ are updated one after the other using gradient descent steps
along their respective gradients:
\begin{align}
  \grad_\theta L_{\text{gen}}(\theta, \phi) &= \grad_\theta h(p_{\theta},a_\phi), \label{eq:generator-grad} \\
  \grad_\phi L_{\text{adv}}(\theta, \phi) &= -\grad_\phi h(p_{\theta},a_\phi). \label{eq:proxy-grad}
\end{align}

\subsection{GANs minimize divergences} \label{sec:gan-objectives}

A common theme in the GAN literature is analysis based on what we call the
\textit{optimal adversary assumption}. This is the assumption that, before each
generator update, we have found the adversary $a_\phi$ which maximizes
$h(p_\theta, a_\phi)$ given the current value of $\theta$.
To be precise, we define a class of permissible adversary functions $\gF$. This
is often simply the space of all functions mapping $\gX\rightarrow \gA$
\citep{goodfellow2014generative}, but is in some GAN variants constrained by,
e.g., a Lipschitz constant~\citep{arjovsky2017wasserstein}. Then we call the
adversary $a_\phi$ optimal for a particular value of $\theta$ if and only if
$h(p_\theta, a_\phi) = \max_{a \in \gF}h(p_{\theta},a)$.

In practice, the neural network $a_\phi$ cannot represent every $a \in \gF$ and
so it may not be able to parameterize an optimal adversary for a given $\theta$.
As is common in the literature, we assume that the neural network is expressive
enough that this is not an issue, i.e., we assume that for any $\theta$, there
exists at least one $\phi\in\Phi$ resulting in an optimal adversary. Then,
noting that there may be multiple such $\phi\in\Phi$, we define $\Phi^*(\theta)$
to be the set of all optimal adversary parameters. That is, $\Phi^*(\theta) =
\{\phi \in \Phi \mid h(p_\theta, a_\phi) = \max_{a \in \gF}h(p_{\theta},a) \}$
and the optimal adversary assumptions says that before each update of $\theta$
we have found $\phi \in \Phi^*(\theta)$. We emphasize that, in part due to the
limited number of gradient updates performed on $\phi$, this assumption is
essentially never true in practice. This paper presents a method to improve the
training of GANs by addressing this issue.

The optimal adversary assumption simplifies analysis of GAN training
considerably. Instead of being a two-player game, it turns into a case of
minimizing an objective with respect to $\theta$ alone. We denote this
objective
\begin{equation}\label{eq:define-m}
  \gM(p_{\theta}) = \max_{a \in \gF}h(p_{\theta},a) = h(p_\theta, a_{\phi^*}) \quad \text{where} \quad \phi^*\in\Phi^*(\theta).
\end{equation}
For example, \citet{goodfellow2014generative} showed that using the objective
presented in \cref{eq:original-h} results in $\gM(p_{\theta}) =
2\cdot\text{JSD}(\ptrue || p_\theta) - \log 4$, where $\text{JSD}$ is the
Jensen-Shannon divergence. By making the optimal adversary assumption, they
could prove that their GAN training procedure would converge, and would minimize
the Jensen-Shannon divergence between $\ptrue$ and $p_\theta$.

A spate of research following the introduction of the original GAN objective has
similarly made use of the optimal adversary assumption to propose GANs which
minimize different divergences. For example, Wasserstein GANs
(WGANs)~\citep{arjovsky2017wasserstein} minimize a Wasserstein distance. MMD
GANs~\citep{li2017mmd} minimize a distance known as the maximum mean
discrepancy. \citet{nowozin2016fgan} introduce f-GANs which minimize
f-divergences, a class including the Kullback-Leibler and Jensen-Shannon
divergences. We emphasize that this is by no means an exhaustive list. Like
these studies, this paper is motivated by the perspective that, under the
optimal adversary assumption, GANs minimize a divergence. However, the GAN
framework can also be viewed from a more game-theoretic
perspective~\citep{kodali2017convergence,grnarova2018online}.

\section{Does an optimal adversary lead to optimal gradients?}

As introduced above, the training of an adversary does not need to be considered
in any analysis if it is simply assumed to always be optimal. From this
perspective, the goal of training GANs can be seen as learning the generator to
minimize $\gM(p_{\theta})$. This leads to the question: assuming that we have an
optimal adversary, can we compute the gradient required for the generator
update, $\grad_\theta \gM(p_{\theta})$? To clarify, assume that we have
generator parameters $\theta'$, and have found $\phi^*\in\Phi^*(\theta')$ such
that $h (p_{\theta'}, a_{\phi^*})$ and $\gM(p_{\theta'})$ are equal in value. We
then want to take a gradient step on $\theta'$ to minimize $\gM(p_{\theta'})$.
Virtually all GAN methods do this by assuming that $\gM(p_{\theta'})$ and $h
(p_{\theta'}, a_{\phi^*})$ have equal gradients with respect to $\theta$ at
$\theta'$. That is, it is assumed that $\grad_\theta
\gM(p_{\theta})\mid_{\theta=\theta'}$ is equal to the \textit{partial
  derivative}\footnote{We use $D_1 h (p_\theta, a_\phi)$ to denote the partial
  derivative of $h(p_\theta, a_\phi)$ with respect to $\theta$ with $\phi$ kept
  constant. Similarly, we will use $D_2 h (p_\theta, a_\phi)$ to denote
  the derivative of $h (p_\theta, a_\phi)$ with respect to $\phi$, with $\theta$
  held constant.} $D_1 h (p_\theta, a_{\phi^*})\mid_{\theta=\theta'}$. It is not
immediately obvious that this is true.

In the GAN literature this concern has largely been overlooked, with a few
treatments for specific GAN types, see
e.g.~\citet{arjovsky2017wasserstein,goodfellow2014generative}. In particular,
\citet{arjovsky2017wasserstein} invoke (but do not explicitly prove) an
extension of \cref{theorem:1} in \citet{milgrom2002envelope} to prove that the
Wasserstein GAN has optimal gradients if the adversary is optimal, i.e. $D_1 h
(p_\theta, a_{\phi^*})\mid_{\theta=\theta'} = \grad_\theta
\gM(p_{\theta})\mid_{\theta=\theta'}$. We note that this extension can, in fact,
be used to prove that GANs in general have this property under fairly weak
assumptions:
\begin{theorem}\label{theorem:1}
  Let $\gM(p_{\theta})=h(p_{\theta},a_{\phi^*})$ for any
  $\phi^*\in\Phi^{*}(\theta)$, as defined in \cref{eq:define-m}. Assuming that
  $\gM(p_{\theta})$ is differentiable \wrt $\theta$ and
  $h(p_{\theta},a_{\phi})$ is differentiable \wrt $\theta$ for all
  $\phi\in\Phi^{*}(\theta)$, then if $\phi^{*}\in\Phi^{*}(\theta)$ we have
  \begin{equation}
    \label{eq:equal-grad}
    \grad_{\theta}\gM(p_{\theta})=D_1 h(p_{\theta},a_{\phi^{*}}).
  \end{equation}
\end{theorem}
See \cref{proof-theo-1} for a proof. We emphasize \cref{theorem:1} applies only
if the adversary is optimal. If this is not the case we cannot quantify, and so
cannot directly minimize or account for, the discrepancy between
$\grad_{\theta}\gM(p_{\theta})$ and $D_1 h(p_{\theta},a_{\phi^{*}})$. Instead of
attempting to do so, we consider an approach that drives the parameters towards
regions where $\phi^{*}\in\Phi^{*}(\theta)$ so that \cref{theorem:1} can be
invoked.

\subsection{Adversary constructors}

To see how we may impose the constraint that \cref{eq:equal-grad} is true, we
consider a trivial relationship between any generator and the corresponding
optimal adversary. If an optimal adversary exists for every $\theta\in\Theta$
then there exists some, possibly non-unique, function $f:\Theta\rightarrow\Phi$
that maps from any generator to a corresponding optimal adversary. That is, for
all $\theta \in \Theta$, $f(\theta) = \phi^*\in\Phi^{*}(\theta)$ in which case
$h(p_{\theta},a_{f(\theta)})=\max_{a\in\gF}h(p_{\theta},a)$. We refer to such a
function as an \textit{adversary constructor}. In an ideal scenario, we could
compute the output of an adversary constructor, $f(\theta)$, for any $\theta$.
We could then invoke \cref{theorem:1} and the generator could be updated with
the gradient $\grad_{\theta}\gM(p_{\theta})=D_{1}h(p_{\theta},a_{f(\theta)})$.
In practice, computing $f(\theta)$ is infeasible and we can only approximate the
optimal adversary parameters with gradient descent. There is therefore a
mismatch between GAN theory, where \cref{theorem:1} is often invoked, and
practice, where the conditions to invoke it are essentially never satisfied. How
then, can we address this problem? We look to the adversary constructors, which
provide a condition that must be satisfied for the optimal adversary assumption
to be true. Adversary constructors allow us to account for the influence of
$\theta$ on $\phi$ by considering the total derivative
$\grad_\theta h(p_{\theta},a_{f(\theta)})$. We prove in \cref{proof-cor-1} that
a comparison with the result of \cref{theorem:1} leads to \cref{cor:1}. In the
next section, we motivate \advas{} as an attempt to fulfill a condition
suggested by this corollary.
\begin{corollary} \label{cor:1} Let $f:\Theta\rightarrow\Phi$ be a
  differentiable mapping such that for all $\theta \in \Theta$,
  $\gM(p_{\theta})=h(p_{\theta},a_{f(\theta)})$. If the conditions in
  \cref{theorem:1} are satisfied and the Jacobian matrix of $f$ with respect to
  $\theta$, $\mJ_{\theta}(f)$ exists for all $\theta\in\Theta$ then
  \begin{equation}\label{eq:zero-tot}
    \trans{D_2 h(p_{\theta},a_{f(\theta)})} \mJ_{\theta}(f) = 0.
  \end{equation}
\end{corollary}

\section{Assisting the adversary}
\label{sec:advas}

\cref{cor:1} tells us that $\trans{D_2 h(p_{\theta},a_{\phi})} \mJ_{\theta}(f)$
will be zero whenever \cref{theorem:1} can be invoked. This makes
\cref{eq:zero-tot} a necessary, but not sufficient, condition for the invocation
of \cref{theorem:1}. This suggests that the magnitude of
$\trans{D_2 h(p_{\theta},a_{\phi})} \mJ_{\theta}(f)$ could be a measure of how
``close'' $D_1 h(p_{\theta},a_{\phi^{*}})$ is to the desired gradient
$\grad_{\theta}\gM(p_{\theta})$. However, the Jacobian $\mJ_{\theta}(f)$ is not
tractable so $\trans{D_2 h(p_{\theta},a_{\phi})} \mJ_{\theta}(f)$ cannot be
computed. The only term we can calculate in practice is
$D_2 h(p_{\theta},a_{\phi})$, exactly the gradient used to train the adversary.
If $D_2 h(p_{\theta},a_{\phi})$ is zero, then
$\trans{D_2 h(p_{\theta},a_{\phi})} \mJ_{\theta}(f)$ is zero. The magnitude of
$D_2 h(p_{\theta},a_{\phi})$ could therefore be an approximate measure of
``closeness'' instead of $\trans{D_2 h(p_{\theta},a_{\phi})} \mJ_{\theta}(f)$.
This leads to an augmented generator loss, which regularizes generator updates
to reduce the magnitude of $D_2 h(p_{\theta},a_{\phi})$. It has a scalar
hyperparameter $\lambda \geq 0$, but \cref{sec:normalize} provides a heuristic
which can remove the need to set this hyperparameter.
\begin{align}
  \label{eq:advas-objective}
  L^{\mr{\advas}}_{\mr{gen}}(\theta,\phi) &= L_{\mr{gen}}(\theta,\phi) + \lambda\madvas{},\\
  \intertext{with}
  \madvas{} &= \norm{\grad_\phi {L_{\mr{adv}}}(\theta,\phi)}_2^2, \label{eq:advas-def}
\end{align}
recalling that
$\grad_\phi {L_{\mr{adv}}}(\theta,\phi) = -D_2 h(p_{\theta},a_{\phi}) $. We
emphasize that $\madvas{}$ is the same as that found in previous
work~\citep{mescheder2017numerics,nagarajan2017gradient}.

Figuratively, \advas{} changes the generator updates to move in a conservative
direction that does not over-exploit the adversary's sub-optimality. Consider
the generator and adversary as two players attempting to out-maneuver one
another. From \cref{eq:general-form}, we see that the generator should learn to
counteract the best possible adversary, rather than the current adversary. If
the current adversary is sub-optimal, allowing it to catch up would yield better
updates to the generator. One way to achieve this is to update the generator in
a way that helps make the current adversary optimal. This behavior is exactly
what \advas{} encourages. In this sense, it assists the adversary, leading to
it's name, the Adversary's Assistant. We emphasize that using \advas{} involves
making only a small modification to a GAN training algorithm, but for
completeness we include pseudocode in \cref{sec:algorithm}.

\subsection{\advas{} preserves convergence results}
\label{sec:preserve}

\advas{} has several desirable properties which support its use as a
regularizer: (1) it does not interfere with the update on $\phi$, and recall
that perfectly optimizing $\phi$ leads to
$h(p_{\theta},a_{\phi})=\gM(p_{\theta})$. (2) Under mild conditions, $\grad_\phi
r(\theta;\phi) \mid_{\phi=\phi^*} $ is zero for an optimal ${\phi^*}$ and so
$\grad_{\theta}\lgenadvas(\theta, \phi^*)=\grad_{\theta}\gM(p_{\theta})$. These
properties imply that, under the optimal adversary assumption, optimizing
$\lgenadvas$ is in fact equivalent to optimizing $\lgen$. See
\cref{sec:preserve-fixed-points} for a proof. Therefore any convergence analysis
which relies on the optimal adversary assumption is equally applicable when
\advas{} is included in the loss. Regarding the mild conditions in property (2),
we require that $\phi^*$ be a stationary point of $h(p_\theta, a_{\phi})$. This
is true as long as $h$ is differentiable \wrt $\phi$ at $(\theta, \phi^*)$ and
$\phi^*$ does not lie on a boundary of $\Phi$. The optimal adversary parameters,
$\phi^*$, cannot lie on a boundary unless, for example, weight clipping is used
as in \citet{arjovsky2017wasserstein}. In such cases, we cannot speak to the
efficacy of applying \advas{}.

We make the additional observation that for some GAN objectives, minimizing
$\madvas$ alone (as opposed to $L_{\text{gen}}$ or $L^{\mr{\advas}}_{\mr{gen}}$)
may match $p_\theta$ and $\ptrue{}$. We show this in \cref{sec:proof-when-advas}
for the WGAN objective~\citep{arjovsky2017wasserstein}. In particular, for all
$\phi\in\Phi$, $\madvas$ is zero and at a global minimum whenever
$p_\theta=\ptrue{}$. Experimental results in \cref{sec:results-advas-alone}
support this observation. However, the results appear worse than those obtained
by optimizing either $L_{\text{gen}}$ or $L^{\mr{\advas}}_{\mr{gen}}$.

\begin{figure}[t]
  \centering
  \includegraphics[scale=1]{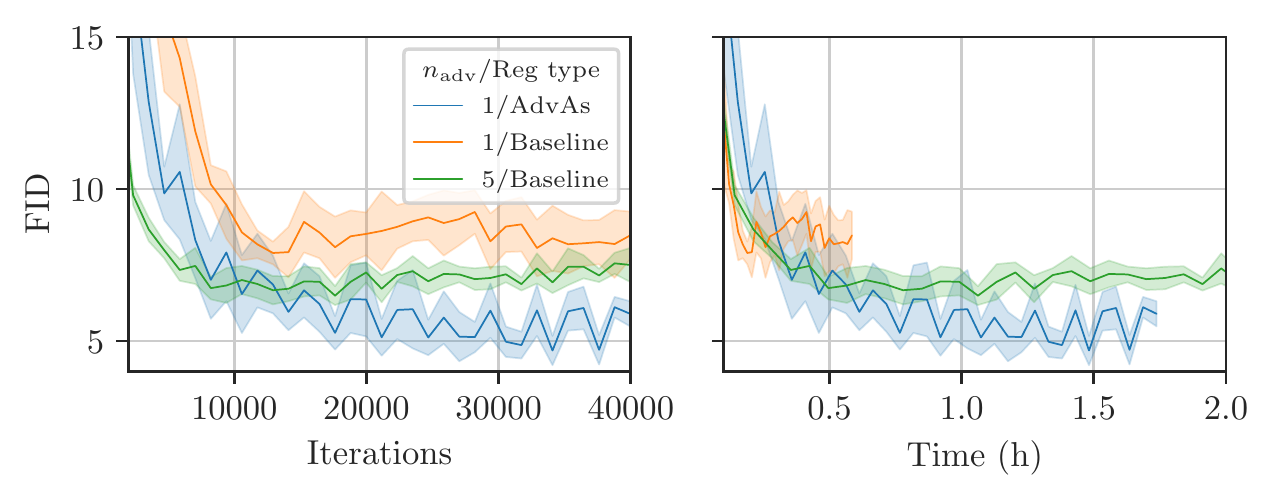}
  \caption{FID scores throughout training for the WGAN-GP objective on MNIST,
    estimated using 60\,000 samples from the generator. We plot up to a maximum
    of 40\,000 iterations. When plotting against time (right), this means some
    lines end before the two hours we show. The blue line shows the results with
    \advas{}, while the others are baselines with different values of $\nadv$. }
  \label{fig:mnist-wgan-fid1000}
\end{figure}
\subsection{Estimating the \advas{} loss}\label{sec:estimate}
It is not always possible, and seldom computationally feasible, to compute the
\advas{} regularization term $\madvas{}$ exactly. We instead use a stochastic
estimate. This is computed by simply estimating the gradient $\grad_\phi
{L_{\mr{adv}}}(\theta,\phi)$ with a minibatch and then taking the squared
L2-norm of this gradient estimate. That is, defining
$\tilde{L}_{\mr{adv}}(\theta,\phi)$ as an unbiased estimate of the adversary's
loss, we estimate $\madvas{}$ with

\begin{equation}
  \tilde{r}(\theta, \phi) = \norm{\grad_\phi {\tilde{L}_{\mr{adv}}}(\theta,\phi)}_2^2.
\end{equation}

Although the gradient estimate is unbiased, taking the norm results in a biased
estimate of $\madvas{}$. However, comparisons with a more computationally
expensive unbiased estimate\footnote{Computing an unbiased estimate can be done
  using the following: consider two independent and unbiased estimates of $\grad_\phi {L_{\mr{adv}}}(\theta,\phi)$ denoted $\XXX,\XXX'$. Then
  $\mean{\XXXt\XXX'}=\trans{\mean{\XXX}}\mean{\XXX'}=\norm{\grad_{\phi}L_{\mr{adv}}(\theta,\phi)}_2^2$. This implies that multiplying two estimates using independent samples is unbiased.} did not reveal a significant
difference in performance.

\subsection{Removing the hyperparameter $\lambda$}\label{sec:normalize}
\begin{wrapfigure}[25]{r}{.5\textwidth}
  \vspace{-1.2em}
  \begin{minipage}[t]{0.5\textwidth}
    \captionof{table}{FID and IS scores on CIFAR10 using AutoGAN with and
      without \advas{}.}
      \begin{tabular}{lll}
        \toprule
        & IS $\pm\sigma$ & FID $\pm\sigma$ \\
        \midrule
        \advas{} ($\nadv=2$) & $8.4\pm 0.1$ & $14.5\pm 1.0$ \\
        Baseline ($\nadv=5$) & $8.3\pm 0.1$ & $15.0\pm 0.7$ \\
        \bottomrule
      \end{tabular}
      \label{tab:cifar-fid50000}
  \end{minipage}
  \vfill
  \begin{minipage}{0.5\textwidth}
    \centering
      \includegraphics{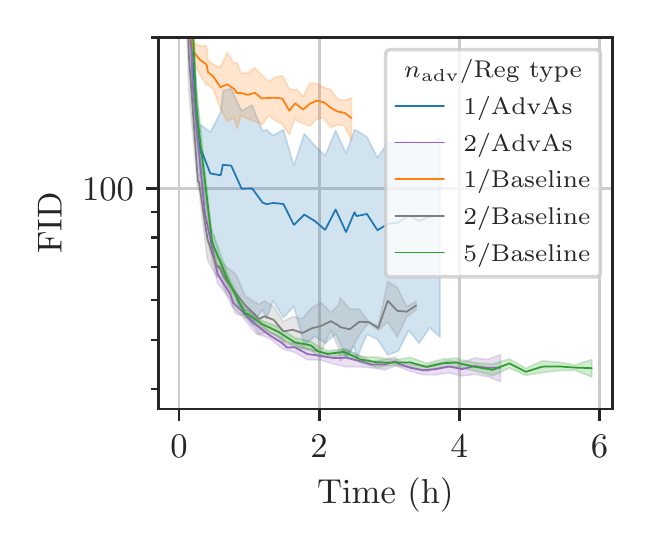}
      \captionof{figure}{FID scores on CIFAR10 using AutoGAN from baselines and
        \advas{} plotted with a log y-axis against running time for different
        values of $\nadv$. We see that \advas{} with $\nadv=2$ yields the lowest
        FID scores at every point during training.}
      \label{fig:cifar-fid1000}
  \end{minipage}
\end{wrapfigure} \cref{eq:advas-objective} introduces a hyperparameter, $\lambda$, which we would prefer not to perform a grid-search on. Setting
$\lambda$ to be too great can destabilize training. Conversely, setting it to be
too small gives similar results to not using \advas{}. We therefore introduce a
heuristic which can be used to avoid setting the hyperparameter. Our experiments
suggests that this is often a good choice, although manually tuning $\lambda$
may yield greater gains. This heuristic involves considering the magnitudes of
three gradients, and so we first define the notation,
\begin{align*}
  \gorig{}(\theta, \phi) &= \grad_{\theta}L_{\mr{gen}}(\theta,\phi), \\
  \gadvas{}(\theta, \phi) &= \grad_{\theta}\tilde{r}(\theta,\phi), \\
  \gtotal{}(\theta, \phi, \lambda) &= \grad_{\theta}L^{\mr{\advas}}_{\mr{gen}}(\theta,\phi) \\
                         &= \vg_{\mr{orig}}(\theta, \phi) + \lambda \vg_{\mr{\advas{}}}(\theta, \phi).
\end{align*}
The heuristic can be interpreted as choosing $\lambda$ at each iteration to
prevent the total gradient, $\gtotal{}(\theta,\phi,\lambda)$, from being
dominated by the \advas{} term. Specifically, we ensure the magnitude of
$\lambda \gadvas{}(\theta, \phi)$ is less than or equal to the magnitude of
$\gorig{}$ by setting

\begin{equation}
  \lambda = \min \left( 1, \frac{\norm{\gorig{}(\theta, \phi)}_2}{\norm{\gadvas{}(\theta, \phi)}_2}\right)
\end{equation}

at every iteration. We then perform gradient descent along $\gtotal{}(\theta,
\phi, \lambda)$. This technique ensures that $\lambda$ is bounded above by $1$.

\section{Experiments}
\label{sec:experiments}

We demonstrate the effect of incorporating \advas{} into GAN training using
several GAN architectures, objectives, and datasets. Our experiments complement
those of \citet{nagarajan2017gradient} and \citet{mescheder2017numerics}. In
each case, we compare GANs trained with \advas{} with baselines that do not use
\advas{} but are otherwise identical. We first demonstrate the use of \advas{}
in conjunction with the \wgan{} objective~\citep{gulrajani2017improved} to model
MNIST~\citep{lecun1998gradientbased}. In this experiment, we compare the
performance gains achieved by \advas{} to a reasonable upper bound on the gains
achievable with this type of regularization. We further support these findings
with experiments on CIFAR10~\citep{krizhevsky2009learning} using
AutoGAN~\citep{gong2019autogan}, an architecture found through neural
architecture search. We then demonstrate that \advas{} can improve training on
larger images using StyleGAN2~\citep{karras2020analyzing} on
CelebA~\citep{liu2015deep}. We quantify each network's progress throughout
training using the FID score~\citep{heusel2017gans}. Since \advas{} increases
the computation time per iteration, we plot training progress against time for
each experiment. We also present inception scores
(IS)~\citep{salimans2016improved} where applicable. We estimate scores in each
case with 5 random seeds and report the standard deviation ($\sigma$) as a
measure of uncertainty.

\advas{} aims to improve performance by coming closer to having an optimal
adversary. Another common way to achieve this is to use a larger number of
adversary updates ($\nadv$) before each generator update. For each experiment,
we show baselines with the value of $\nadv$ suggested in the literature. Noting
that the computational complexity is $\gO(\nadv)$ and so keeping $\nadv$ low is
desirable, we find that \advas{} can work well with lower values of $\nadv$ than
the baseline. For a fair comparison, we also report baselines trained with these
values of $\nadv$.

For MNIST and CelebA, we avoid setting the hyperparameter $\lambda$ by using the
heuristic proposed in \cref{sec:normalize}. We found for CIFAR10 that manually
tuning $\lambda$ gave better performance, and so set $\lambda = 0.01$.
Additionally, on MNIST and CelebA, the methods we consider use regularization in
the form of a gradient penalty~\citep{gulrajani2017improved} for training the
adversary. This is equivalent to including a regularization term
$\gamma_{\mr{adv}}(\phi)$ in the definition of $\ladv{}$. That is,
$\ladv{}(\theta, \phi) = -\grad_\phi h(p_{\theta},a_\phi) +
\gamma_{\mr{adv}}(\phi)$. Following \cref{eq:advas-def} this regularization term
is included in the \advas{} term $\madvas{}$. Another practical detail is that
AutoGAN and StyleGAN2 are trained with a hinge loss~\citep{lim2017geometric}.
That is, when computing the adversary's loss $L_{\text{adv}}(\theta, \phi)$, its
output $a_\phi(\vx)$ is truncated to be below $+1$ for real images, or above
$-1$ for generated images. This prevents it from receiving gradient feedback
when its predictions are both accurate and confident. Intuitively, this stops
its outputs becoming too large and damaging the generator's training. However,
this truncation is not present when updating the generator. This means that the
generator minimizes a different objective to the one maximized by the adversary,
and so it is not exactly a minimax game. It is not clear that it is beneficial
to calculate the \advas{} regularization term using this truncation. We found
that better performance was obtained by computing $\madvas{}$ without
truncation, and do this in the reported experiments.


\subsection{WGAN-GP on MNIST} \label{sec:mnist}

We use a simple neural architecture: the generator consists of a fully-connected
layer followed by two transposed convolutions. The adversary has three
convolutional layers. Both use instance
normalization~\citep{ulyanov2017instance} and ReLU non-linearities; see
\cref{sec:experimental-setup} for details. We compare using \advas{} with
$\nadv=1$ against the baseline for $\nadv\in\tub{1,5}$ where $\nadv=5$ is
suggested by \citet{gulrajani2017improved}. \cref{fig:mnist-wgan-fid1000} shows
the FID scores for each method throughout training. We see that using \advas{}
with $\nadv{} = 1$ leads to better performance on convergence; even compared to
the baseline with $\nadv{} = 5$, the best FID score reached is improved by
$28\%$.


\begin{figure}[t]
  \centering
  \includegraphics[scale=1]{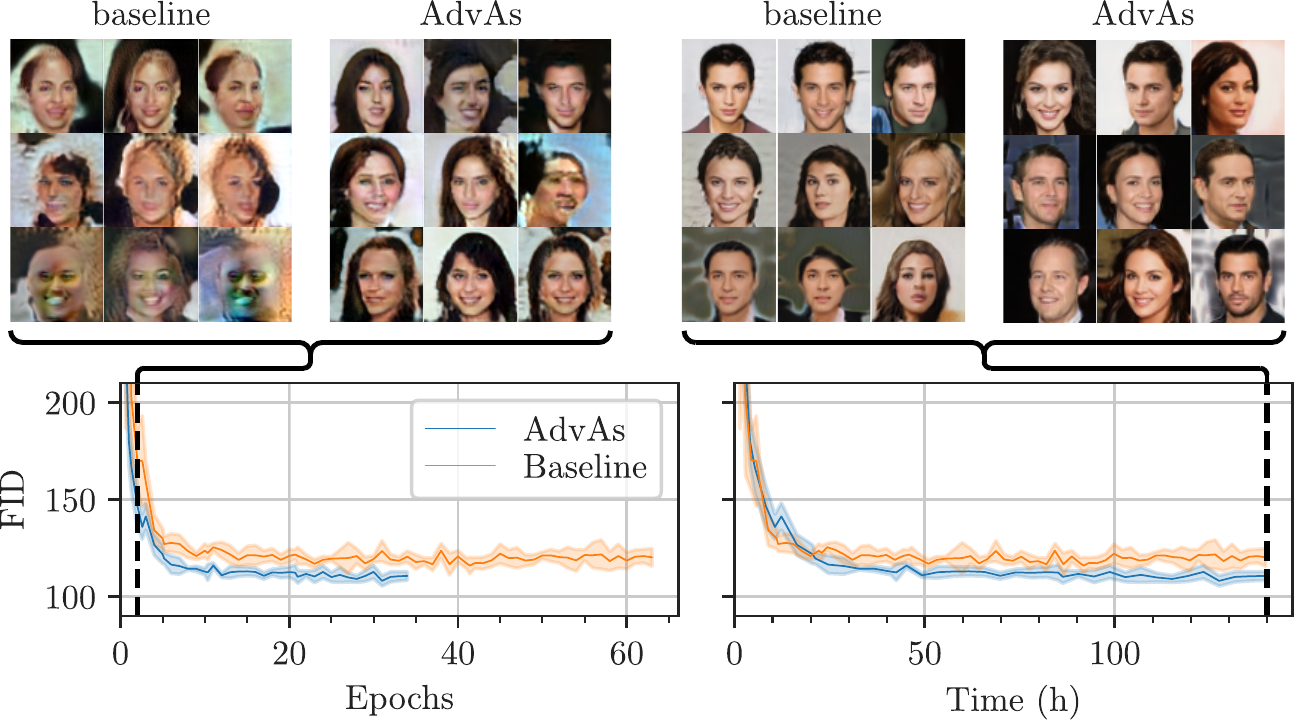}
  \caption{\textbf{Bottom:} FID scores throughout training estimated with $1000$
    samples, plotted against number of epochs (left) and training time (right).
    FID scores for \advas{} decrease more on each iteration at the start of
    training and converge to be $7.5\%$ lower. \textbf{Top:} The left two
    columns show uncurated samples with and without \advas{} after 2 epochs. The
    rightmost two columns show uncurated samples from networks at the end of
    training. In each grid of images, each row is generated by a network with a
    different training seed and shows 3 images generated by passing a different
    random vector through this network. \advas{} leads to obvious qualitative
    improvement early in training.}
  \label{fig:celeba-training}
  \vspace{-1em}
\end{figure}
\subsection{AutoGAN on CIFAR10} \label{sec:cifar}

We next experiment on the generation of CIFAR10~\citep{krizhevsky2009learning}
images. We use AutoGAN~\citep{gong2019autogan}, which has a generator
architecture optimized for CIFAR10 using neural architecture search. It is
trained with a hinge loss, as described previously, an exponential moving
average of generator weights, and typically uses $\nadv = 5$.
\Cref{fig:cifar-fid1000} shows FID scores throughout training for various values
of $\nadv$, with and without \advas{}, each computed with $1000$ samples.
\Cref{tab:cifar-fid50000} shows FID scores at the end of training for the best
performing value of $\nadv$ for each method, estimated with $50\,000$ samples.
For a fixed $\nadv$ of either 1 or 2, using \advas{} improves the FID score. In
fact, with $\nadv = 2$, the performance with \advas{} is indistinguishable from
the baseline with the suggested setting of $\nadv = 5$. Unlike for MNIST,
\advas{} does not outperform the baseline with high enough $\nadv$. We
hypothesize that this is because, with an architecture highly optimized for
$\nadv{} = 5$, the adversary is closer to being optimal when trained with
$\nadv=5$. Assuming this is the case, we would not expect \advas{} to improve
training compared to a baseline with sufficient $\nadv{}$. Still, our results
show that applying \advas{} allows the same performance with a lower $\nadv{}$.

\subsection{StyleGAN2 on CelebA} \label{sec:celeba} To demonstrate that \advas{}
improves state-of-the-art GAN architectures and training procedures, we consider
StyleGAN2~\citep{karras2020analyzing}. We train this as proposed by
\citet{karras2020analyzing} with a WGAN-like objective with gradient
penalty~\citep{gulrajani2017improved}, an exponential moving average of the
generator weights, and various forms of regularization including path length,
R1, and style-mixing regularization. More detail on these can be found in
\citet{karras2020analyzing}, but we merely wish to emphasize that considerable
effort has been put into tuning this training procedure. For this reason, we do
not attempt to further tune $\nadv$, which is $1$ by default. Any improvements
from applying \advas{} indicate a beneficial effect not provided by other forms
of regularization used.

\Cref{fig:celeba-training} compares the training of StyleGAN2 on CelebA at
64$\times$64 resolution with and without the \advas{} regularizer. Using
\advas{} has two main effects: (1) the generated images show bigger improvements
per epoch at the start of training; and (2) the final FID score is improved by
$7.5\%$. Even accounting for its greater time per iteration, the FID scores
achieved by \advas{} overtake the baseline after one day of training. We verify
that the baseline performance is similar to that reported by
\citet{zhou2019hype} with a similar architecture.

\section{Related work}
\label{sec:related}

We motivated \advas{} from the perspective of the optimal adversary assumption.
In this sense, it is similar to a large body of work aiming to improve and
stabilize GAN training by better training the adversary. \advas{} differs
fundamentally due to its focus on the training the generator rather than the
adversary. This other work generally affects the discriminator in one of two
broad ways: weight constraints and gradient penalties~\citet{brock2019large}.
Weight normalization involves directly manipulating the parameters of the
adversary, such as through weight clipping~\citep{arjovsky2017wasserstein} or
spectral normalization~\citep{miyato2018spectral}. Gradient
penalties~\citep{kodali2017convergence,roth2017stabilizing,gulrajani2017improved}
impose soft constraints on the gradients of the adversary's output with respect
to its input. Various forms exist with different motivations; see
\citet{mescheder2018which} for a summary and analysis. \advas{} may appear
similar to a gradient penalty, as it operates on gradients of the adversary.
However, the gradients are \wrt the adversary's parameters rather than its
input. Furthermore, \advas{} is added to the generator's loss and not the
adversary's.

Regularizing generator updates has recently received more attention in the
literature~\citep{chu2020smoothness,zhang19d,brock2019large}.
\citet{chu2020smoothness} show theoretically that the effectiveness of different
forms of regularization for both the generator and adversary is linked to the
smoothness of the objective function. They present a set of conditions on the
generator and adversary that ensure a smooth objective function, which they
argue will stabilize GAN training. However, they leave the imposition of the
required regularization on the generator to future work. \citet{zhang19d} and
\citet{brock2019large} consider applying spectral normalization~
\citep{miyato2018spectral} to the generator, and find empirically that this
improves performance.

\section{Discussion and conclusions}

We have shown that \advas{} addresses the mismatch between theory, where the
adversary is assumed to be trained to optimality, and practice, where this is
never the case. We show improved training across three datasets, architectures,
and GAN objectives, indicating that it successfully reduces this disparity. This
can lead to substantial improvements in final performance. We note that, while
applying \advas{} in preliminary experiments with
BEGAN~\citep{berthelot2017began} and LSGAN~\citep{mao2017least}, we did not
observe either a significant positive effect, or a significant negative effect
other than the increased time per iteration. Nevertheless, \advas{} is simple to
apply and will, in many cases, improve both training speed and final
performance.



\bibliography{adversarys_assistant.bib}
\bibliographystyle{iclr2021_conference}

\newpage
\appendix

\section{The \advas{} algorithm}
\label{sec:algorithm}

\cref{algo:train} provides pseudocode for training a GAN. Using \advas{} involves
estimating an additional loss term for the generator and, to avoid setting the
hyperparameter $\gamma$, performing normalization of the gradient as described
in \cref{sec:normalize}.

\begin{algorithm}[t]
  \DontPrintSemicolon
  \For{number of training iterations} {
    \For{number of adversary update steps} {
      Estimate adversary loss $L_{\text{adv}}(\theta, \phi)$\;
      Update adversary parameters $\phi$ with gradient descent along $\grad_\phi L_{\text{adv}}(\theta,
      \phi)$\;
    }
    Estimate generator loss $L_{\text{gen}}(\theta, \phi) $\;
    \uIf{use \advas{}} {
      \color{blue}
      Estimate \advas{} loss $\madvas{}$\;
      $g = \text{NormalizeGradient}(\grad_\phi L_{\text{gen}}(\theta, \phi) ,
      \grad_\phi \madvas{})$\;
    }\Else{
      \color{red}
      $g = \grad_\phi L_{\text{gen}}(\theta, \phi)$\;
    }
    Update generator parameters $\theta$ with gradient descent along $g$ \;
  }
  \caption{Train a GAN. To use \advas{}, execute the two \textcolor{blue}{blue}
    lines instead of the \textcolor{red}{red} line.}
  \label{algo:train}
\end{algorithm}

\section{Estimator for the \advas{} loss}
\label{sec:app-estimate-advas}
We can derive an alternative unbiased estimate of $\madvas$ to that presented in
\cref{sec:estimate}. To do this, we follow \citet{nowozin2016fgan} and note that
the objective can generally be written in the form
\begin{equation}
  h(\theta, \phi) = \mathbb{E}_{\vx_1 \sim p_\theta, \vx_2 \sim \ptrue{}} [ g(\phi,\xxx_{1},\xxx_{2}) ] .
\end{equation}
Therefore,
\begin{align} \label{eq:norm-expect}
  \madvas{} = \norm{\mean{\grad_\phi g(\phi,\xxx_1, \xxx_2)}}_2^2,
\end{align}
Using $\pp_{i}$ to denote the $i$th partial derivative of $\phi$, we have that,
\begin{align}
  \norm{\mean{\grad_\phi g(\phi,\xxx_1, \xxx_2)}}_2^2 &= \sum_i \mean{\pp_i g(\phi,\xxx_1,\xxx_2)}^2.
\end{align}
As the sum of unbiased estimators is itself unbiased, we need only find an
unbiased estimate of $\mean{\pp_i g(\phi,\xxx_1,\xxx_2)}^2$, which is of
the form $\mean{x}^{2}$. Using the identity
$\mr{Var}(x) = \mean{x^{2}} - \mean{x}^{2}$, we find, using Monte Carlo estimates
of the expectations, that,
\begin{align}
  \mean{\pp_i g(\phi,\xxx_1,\xxx_2)}^2 &= \mr{Var}(\pp_i g(\phi,\xxx_1,\xxx_2)) + \mean{\paren{\pp_i g(\phi,\xxx_1,\xxx_2)}^{2}} \\
                                       &\approx \frac{1}{N-1}\sum_{n=1}^N \paren{\pp_i g(\phi,\xxx^{(n)}_1,\xxx^{(n)}_2)-\overline{\pp_{i}g}}^2 + \frac{1}{N}\sum_{n=1}^N\paren{\pp_i g(\phi,\xxx^{(n)}_1,\xxx^{(n)}_2)}^{2}\\
  & \equiv \tilde{e}_i(\phi),
\end{align}
where $\vx^{(n)}_1 \sim p_\theta, \vx^{(n)}_2 \sim \ptrue{}$ and
$\overline{\pp_{i}g} = \frac{1}{N}\sum_{n=1}^N\pp_i g(\phi,\xxx^{(n)}_1,\xxx^{(n)}_2)$.
As $\tilde{e}(\phi)$ is an unbiased estimator it follows that
\begin{equation} \label{eq:unbiased}
  \sum_i \tilde{e}_i(\phi) \approx \norm{\mean{\grad_\phi g(\phi,\xxx_1, \xxx_2)}}_2^2
\end{equation}
is also an unbiased estimator. However, calculating $\sum_i \tilde{e}_i(\phi)$
is currently inefficient since it involves estimating the variance of gradients,
which is poorly supported by the implementation of automatic differentiation in
e.g. PyTorch. Therefore, we instead use the following biased estimate of
$\madvas$, by directly taking the norm of the Monte Carlo estimate of the
expectation in \cref{eq:norm-expect},
\begin{equation} \label{eq:biased}
  \tilde{r}(\theta,\phi) \approx \norm{ \frac{1}{N} \sum_{n=1}^N  \grad_\phi g(\phi, \vx^{(n)}_1, \vx^{(n)}_2   }_2^2,
\end{equation}
where $\vx^{(n)}_1 \sim p_\theta, \vx^{(n)}_2 \sim \ptrue{}$. We emphasize again
that comparisons between the unbiased estimate in \cref{eq:unbiased} and the
biased estimate in \cref{eq:biased}, did not suggest a significant difference in
performance.

\section{Additional experimental results}

\subsection{CIFAR10} \label{sec:cifar-samples}
\begin{figure}[h]
  \centering
  \includegraphics[width=\textwidth]{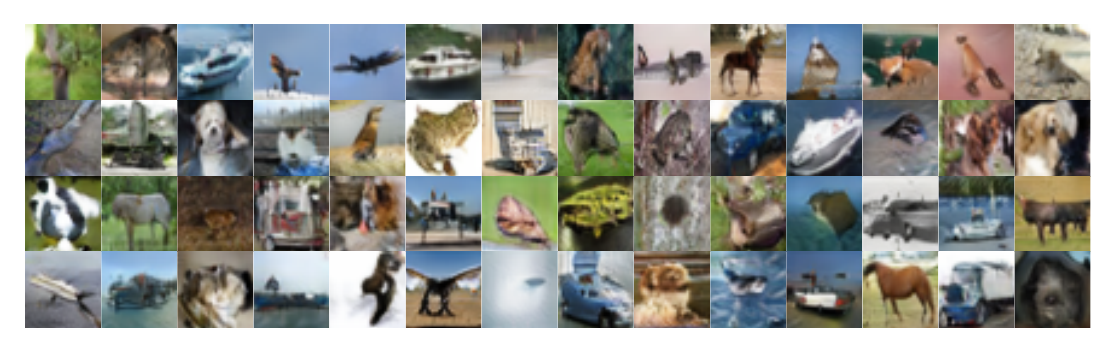}
  \caption{Uncurated samples from the baseline AutoGAN with $\nadv=5$ after
    finishing training.}
  \label{fig:cifar-samples-no-advas}
  \includegraphics[width=\textwidth]{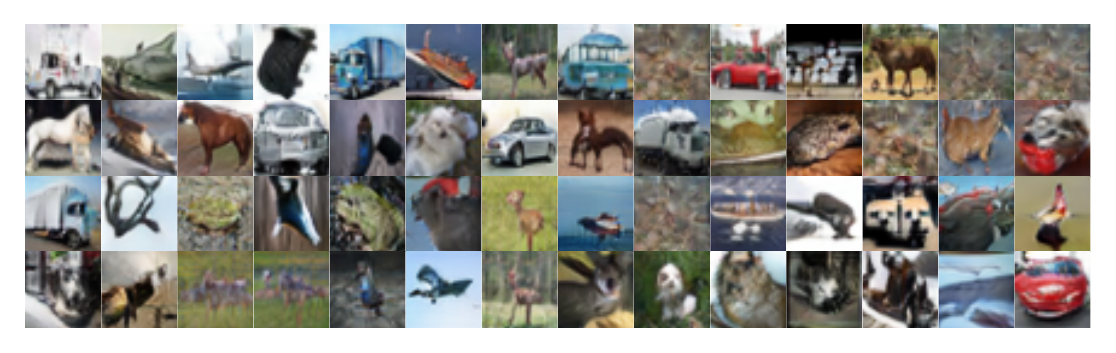}
  \caption{Uncurated samples from the AutoGAN with \advas{} and $\nadv=5$ after
    finishing training.}
  \label{fig:cifar-samples-advas}
\end{figure}

In \cref{fig:cifar-samples-no-advas,fig:cifar-samples-advas}, we present a
random sample of CIFAR10 images from networks trained with the best-performing
values of $\nadv$ with and without \advas{}. The networks further use
exponential moving averages of their weights. Reflecting the similar FID scores
on convergence, there is no obvious qualitative difference between the samples.

\subsection{CelebA} \label{sec:celeba-samples}

\begin{figure}[h]
  \includegraphics[scale=1]{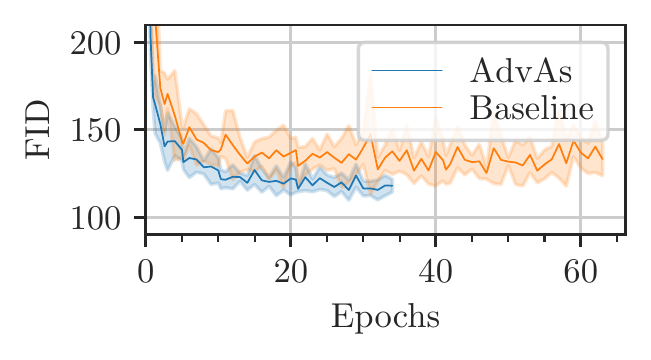}
  \includegraphics[scale=1]{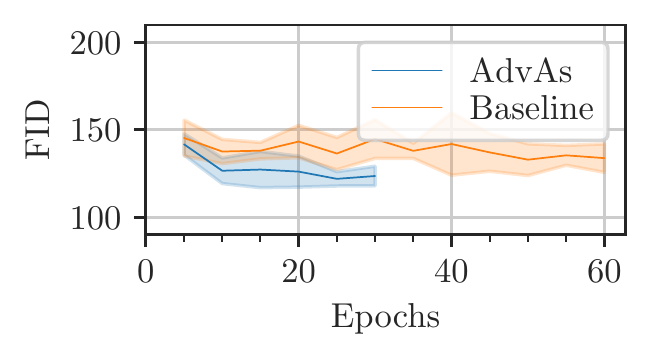}
  \vspace{-1em}
  \caption{FID scores for StyleGAN2 on CelebA-64 calculated with $1000$ samples
    (left), or $202\,599$ (right) without an exponential moving average of the
    weights.}
  \label{fig:celeba-no-ema}
  \includegraphics[scale=1]{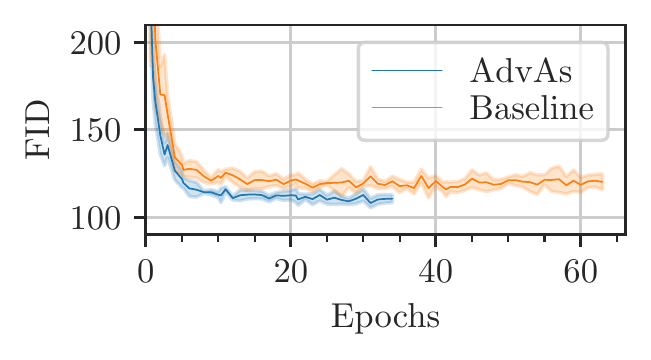}
  \includegraphics[scale=1]{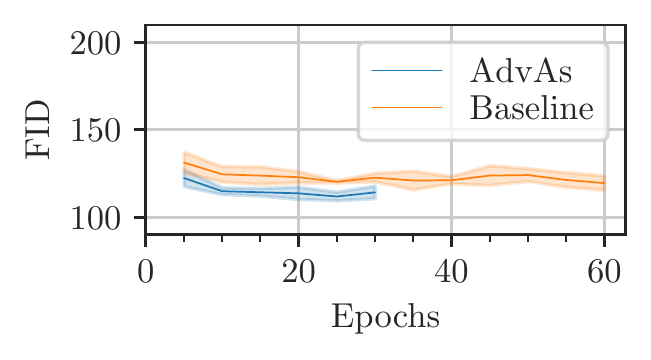}
  \vspace{-1em}
  \caption{FID scores for StyleGAN2 on CelebA-64 calculated with $1000$ samples
    (left), or $202\,599$ (right) with an exponential moving average of the
    weights as in \cref{fig:celeba-training}.}
  \label{fig:celeba-yes-ema}
\end{figure}

\paragraph{Quantitative metrics} In \cref{fig:celeba-training} of the main
paper, we plot FID scores for CelebA throughout training, evaluated with $1000$
samples and an exponential moving average of the generator weights. In
\cref{fig:celeba-no-ema} we show corresponding results without using this
average of generator weights. The results are qualitatively similar, but the FID
scores are slightly higher for each method. Additionally, in the right columns
of \cref{fig:celeba-no-ema,fig:celeba-yes-ema}, we plot a FID score computed
less frequently using 202\,599 samples from the generator; this is as many
samples as there are images in the CelebA dataset.

\begin{figure}[h]
  \vspace{.5cm}
  \centering
  \includegraphics[width=\textwidth]{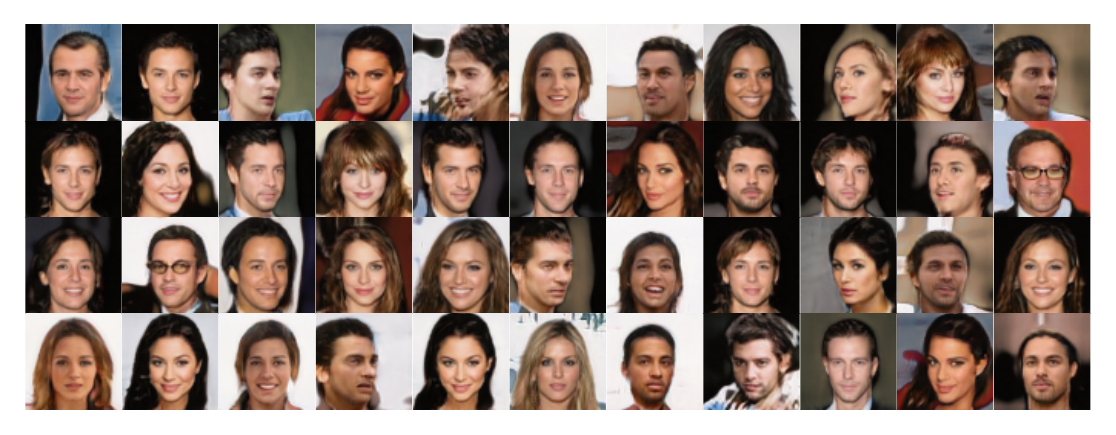}
  \includegraphics[width=\textwidth]{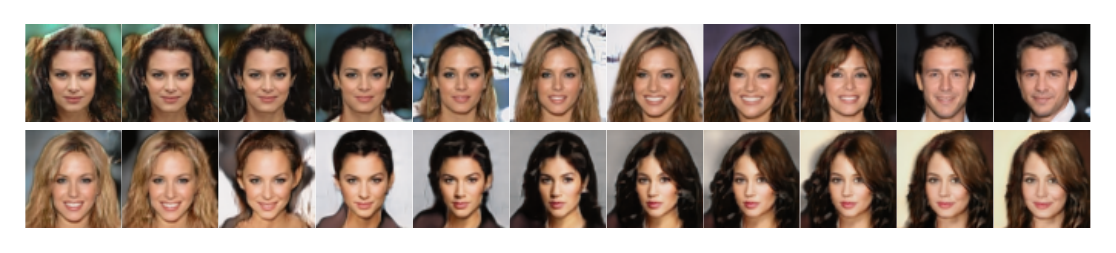}
  \caption{Uncurated samples and interpolations from our baseline StyleGAN2
    trained for 140 hours.}
  \label{fig:samples-no-advas}
  \vspace{1cm}
  \includegraphics[width=\textwidth]{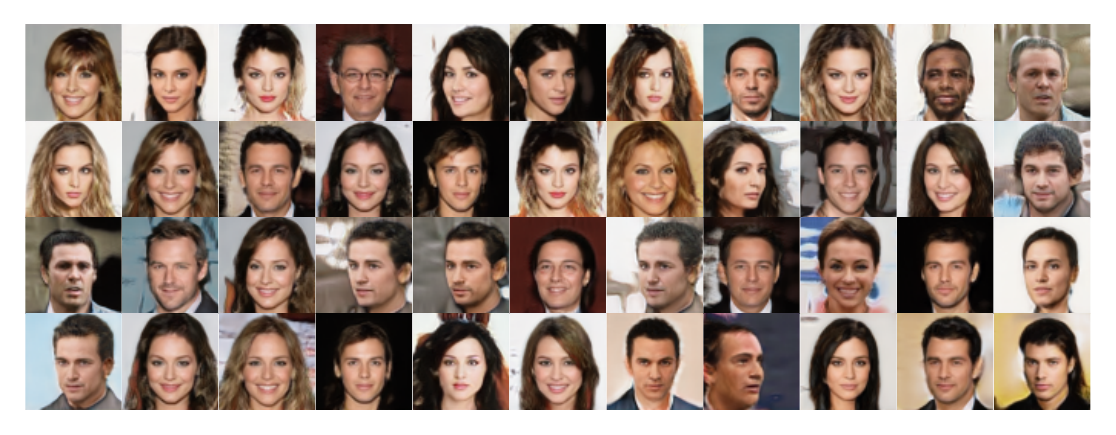}
  \includegraphics[width=\textwidth]{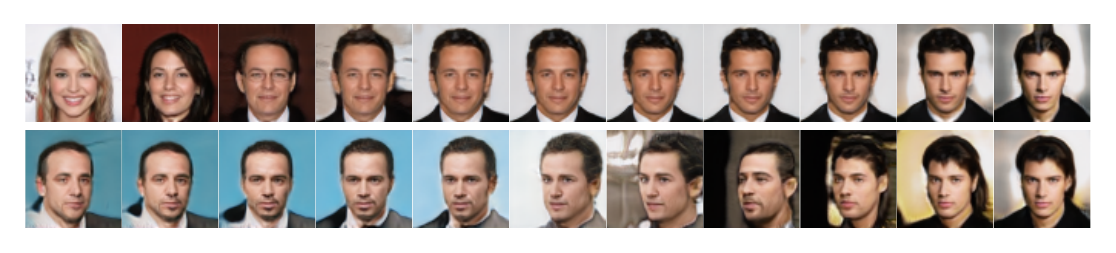}
  \caption{Uncurated samples and interpolations from StyleGAN2 with \advas{}
    trained for 140 hours.}
  \label{fig:samples-advas}
  \vspace{1cm}
\end{figure}

\paragraph{Random samples}
In addition to reporting FID scores for each method in the paper, we provide a
random sample of images from a GAN trained with \advas{}
(\cref{fig:samples-no-advas}), and one trained without
(\cref{fig:samples-advas}). We also show interpolations between randomly sampled
latent variables. If images change smoothly as the latent variables change, this
indicates that the network does not overfit to the training
data~\citep{brock2019large}. The interpolations of networks trained with
\advas{} are at least as smooth as those without.

\subsection{Training with \advas{} alone}\label{sec:results-advas-alone}

\begin{figure}[t]
  \centering
  \includegraphics[width=\textwidth]{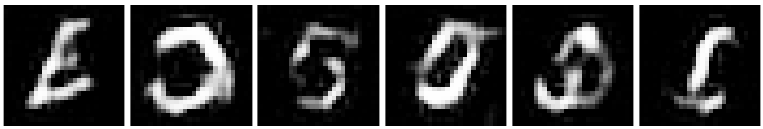}
  \caption{Uncurated samples from a generator trained to minimize $\madvas{}$
    alone. The adversary is trained using its standard loss,
    $L_{\text{adv}}(\theta, \phi)$. We use the WGAN objective (\cref{eq:wgan})
    with weight clipping to enforce the Lipschitz constraint, and train for
    100\,000 iterations. The neural network architecture used was similar to
    that described in \cref{sec:mnist-setup}, but with upsampling rather than
    transposed convolutions, and batch normalization rather than instance
    normalization. The samples exhibit some features of MNIST images, but are
    mostly not recognizable digits. }
  \label{fig:mnist-only-advas}
\end{figure}

As mentioned in \cref{sec:preserve}, for some GAN objectives the \advas{} term
$\madvas{}$ has a global minimum for any value of $\phi$ when
$p_\theta = \ptrue{}$. A natural question is therefore whether the generator can
be fitted to the data purely by minimizing $\madvas{}$.
\Cref{fig:mnist-only-advas} shows that a generator can be learned in this way
for MNIST, although with results that are clearly inferior to minimizing the
standard generator objective. This indicates that, although \advas{} can be used
to directly match the generated and true distributions, the primary beneficial
effect on training comes from using it to ``assist'' an adversary in addition to
using the standard generator loss.

In the figure, the adversary is trained using its standard loss. This is
theoretically unnecessary since setting $p_\theta = \ptrue{}$ will minimise
$\madvas{}$ for any $\phi$. Therefore, $\phi$ could be sampled at initialization
and trained no further. However, we found that training $\phi$ along with
$\theta$ was necessary or the images produced would be qualitatively
significantly worse than those in \cref{fig:mnist-only-advas}. The architecture
used is different to other experiments, using bilinear upsampling rather than
transposed convolutions. This was necessary to achieve the results shown.

\section{Proofs}
\subsection{Proof of \cref{theorem:1}}
\label{proof-theo-1}

This theorem is based on Theorem 1 of \citet{milgrom2002envelope}, which assumed
$\theta \in [0,1]$. We generalize this to $\theta \in \mathbb{R}^n$, as
\citeauthor{milgrom2002envelope} noted was possible but did not explicitly
prove.
The proof is also similar to that of Danskin's
theorem~\citep[p.~717]{bertsekas1997nonlinear}. However, Danskin's theorem makes
different assumptions which do not guarantee that $\grad_\theta\gM(p_\theta)$
exists, and so is in general limited to considering directional derivatives.

Let $\theta, \bar{\theta} \in \mathbb{R}^n$ and $\phi^* \in \Phi^*(\theta)$.
Then, by the definition of $\gM$ in \cref{eq:define-m}, we have that
\begin{align}
  \gM(p_\theta) &= h(p_{\theta},a_{\phi^*}), \\
  \gM(p_{\bar{ \theta }} ) &\geq h(p_{\bar{\theta}},a_{\phi^*}),
\end{align}
and subsequently
\begin{align}
  \gM(p_{\bar{ \theta }} ) - \gM(p_\theta) &\geq h(p_{\bar{\theta}},a_{\phi^*}) - h(p_{\theta},a_{\phi^*}).
\end{align}
We set $\bar{\theta} = \theta + h\cdot\nu$, for some positive scalar $h$ and
$\nu\in\mathbb{R}^n$. Then dividing both sides by $h$ gives
\begin{align}
  \frac{\gM(p_{\theta + h\cdot\nu} ) - \gM(p_\theta)}{h} &\geq \frac{h(p_{\theta + h\cdot\nu},a_{\phi^*}) - h(p_{\theta},a_{\phi^*})}{h}.
\end{align}
Taking the limit of both sides as $h \to 0$ results in directional derivatives in the
direction $\nu$:
\begin{align}
  \grad_\theta^\nu \gM(p_\theta) \geq \grad_\theta^\nu h(p_\theta, a_{\phi^*}).
\end{align}
Due to the Theorem 1's stated restriction that $\grad_{\theta}\gM(p_{\theta})$
and $\grad_{\theta}h(p_{\theta},a_{\phi^{*}})$ both exists, the directional
derivatives are equal to the dot product of $\nu$ and these derivatives:
\begin{align}\label{eq:nu-dot-grad}
  \nu \cdot \grad_\theta \gM(p_\theta) \geq \nu \cdot \grad_\theta h(p_\theta, a_{\phi^*}).
\end{align}
Since this is true for all $\nu \in \mathbb{R}^d$, both of the following must
true for all $u \in \mathbb{R}^d$:
\begin{align}\label{eq:neg-nu-dot-grad}
  u \cdot \grad_\theta \gM(p_\theta) \geq u \cdot \grad_\theta h(p_\theta, a_{\phi^*}), \quad -u \cdot \grad_\theta \gM(p_\theta) \geq -u \cdot \grad_\theta h(p_\theta, a_{\phi^*}).
\end{align}
The only way that these can both be true is if the inequality is in fact an
equality:
\begin{align}
  \nu \cdot \grad_\theta \gM(p_\theta) = \nu \cdot \grad_\theta h(p_\theta, a_{\phi^*}).
\end{align}
Finally, this can only be true for all $\nu \in \mathbb{R}^d$ if the gradients
themselves are equal, leading to the result
\begin{align}
  \grad_\theta \gM(p_\theta) = \grad_\theta h(p_\theta, a_{\phi^*})
\end{align}
and concluding the proof. \qed

\subsection{Proof of \cref{cor:1}}
\label{proof-cor-1}
Let an adversary constructor $f$ be a differentiable function with respect to
$\theta$ such that the \textit{total derivative} of
$h(p_{\theta},a_{f(\theta)})$ is,
\begin{equation}\label{eq:total}
  \grad_\theta h(p_{\theta},a_{f(\theta)}) = D_1 h(p_{\theta},a_{f(\theta)}) + \trans{D_2 h(p_{\theta},a_{f(\theta)})} \mJ_{\theta}(f),
\end{equation}

From the definition of $f$, we have that
$h(p_{\theta},a_{f(\theta)})=\gM(p_{\theta})$ for all $\theta \in \Theta$. From
the conditions of \cref{theorem:1}, $\gM(p_{\theta})$ is differentiable with
respect to $\theta$. Therefore, $\grad_\theta\gM(p_{\theta}) = \grad_\theta
h(p_{\theta},a_{f(\theta)})$. Writing out the total derivative as in
\cref{eq:total} gives
\begin{equation}
  \grad_\theta\gM(p_{\theta}) = D_1 h(p_{\theta},a_{f(\theta)}) + \trans{D_2 h(p_{\theta},a_{f(\theta)})} \mJ_{\theta}(f).
\end{equation}
\cref{theorem:1} says that $\grad_\theta\gM(p_{\theta}) = D_1
h(p_{\theta},a_{f(\theta)})$. Subtracting this from both sides of the above
equation leads to the stated result. \qed

\subsection{Proof that minimizing $\madvas{}$ reaches optimum for WGAN}
\label{sec:proof-when-advas}

To prove that the assumption is correct for the WGAN, we consider its objective
\begin{equation}
  \label{eq:wgan}
  h_{\mr{WGAN}}(p_{\theta},a_{\phi})=\meanp{\vx\sim \ptrue{}}{a_{\phi}(\vx)} - \meanp{\vx\sim p_{\theta}}{a_{\phi}(\vx)}, \quad \text{s.t. } \norm{a_{\phi}}_{L}\leq 1.
\end{equation}
Consider $p_{\theta}=\ptrue{}$. Clearly, this will cause
$h_{\mr{WGAN}}(p_{\theta},a_{\phi})$ to be zero for any adversary $a_\phi$. In
turn, this means that any adversary is optimal since $\grad_\phi
{L_{\mr{adv}}}(\theta,\phi) =0$ and so $\madvas=0$ for all $\phi\in\Phi$. For
any $\phi$, this is a global minimum of $\madvas{}$ with respect to $\theta$.
This indicates that \advas{} promotes $p_{\theta}$ to be equal to $\ptrue{}$
even when the adversary is sub-optimal. Exactly the same analysis can be applied
to various GANs with objectives that can be written similarly as a difference
between expectations of the adversary's
output~\citep{li2015generative,unterthiner2017coulomb}.

\subsection{Proof that \advas{} preserves convergence
  results} \label{sec:preserve-fixed-points}

\textbf{Proposition.} Making the optimal adversary assumption
(\cref{sec:gan-objectives}), $\phi$ is updated at each training iteration to be
a member of $\Phi^*(\theta)$, resulting in an optimal adversary. Then, whether
$\theta$ is updated to decrease $L^{\mr{\advas}}_{\mr{gen}}(\theta,\phi)$ or
$L_{\mr{gen}}(\theta,\phi)$, the fixed points to which the generator can
converge are the same.

\textit{Proof.} Having made the optimal adversary assumption, these two
different generator losses have the gradients
\begin{align}
  \grad_\theta L_{\mr{gen}}(\theta,\phi) &= \grad_{\theta}\gM(p_{\theta}) \\
  \intertext{and}
  \grad_\theta L^{\mr{\advas}}_{\mr{gen}}(\theta,\phi) &= \grad_{\theta}\gM(p_{\theta}) + \lambda \cdot \grad_\theta \madvas{} \quad \text{where} \quad \phi \in \Phi^*(\theta). \label{eq:grad-loss-advas} \\
                                         &= \grad_{\theta}\gM(p_{\theta}) + \lambda \cdot \grad_\theta \norm{\grad_\phi {L_{\mr{adv}}}(\theta,\phi)}_2^2 \\
                                                       &= \grad_{\theta}\gM(p_{\theta}) + \lambda \cdot 2 \big( \underbrace{ \grad_\phi {L_{\mr{adv}}}(\theta,\phi) }_\text{adversary's gradient} \big) \cdot ( \grad_\theta \grad_\phi {L_{\mr{adv}}}(\theta,\phi) \big). \label{eq:grad-loss-advas-expanded}
\end{align}
As long as $\grad_\theta \grad_\phi {L_{\mr{adv}}}(\theta,\phi)$ exists, a
sufficient condition for the gradients of $\grad_\theta
L_{\mr{gen}}(\theta,\phi)$ and $\grad_\theta
L^{\mr{\advas}}_{\mr{gen}}(\theta,\phi)$ to be equal is therefore if the
adversary's gradient term marked in \cref{eq:grad-loss-advas-expanded} is
$\vzero{}$. We will now show that this is the case. To do so, we invoke Fermat's
theorem on stationary points, which states that at least one of the following is
true for each $\phi' \in \Phi^*(\theta)$:
\begin{enumerate}
\item $\phi'$ is in the boundary of $\Phi$. \label{enum:boundary}
\item $h(p_\theta, a_{\phi'})$ is not differentiable at $\phi'$. \label{enum:no-deriv}
\item $\grad_{\phi'} {L_{\mr{adv}}}(\theta,\phi') = \vzero{}$. \label{enum:zero-deriv}
\end{enumerate}
As long as items \ref{enum:boundary} and \ref{enum:no-deriv} are not the case,
we have that $\grad_\phi {L_{\mr{adv}}}(\theta,\phi) = \vzero{}$. Therefore,
\begin{align}
  \grad_\theta L^{\mr{\advas{}}}_{\mr{gen}}(\theta,\phi) &= \grad_{\theta}\gM(p_{\theta}).
\end{align}
Since, under the stated assumptions, both losses have the same gradients with
respect to $\theta$, the sets of fixed points to which the parameters can
converge under each are the same. \qed

\section{Experimental setup} \label{sec:experimental-setup}

\subsection{MNIST} \label{sec:mnist-setup}
We normalize the data to lie between 0 and 1. The generator we use takes as
input a random, unit Gaussian-distributed, vector of dimension 128. This is
mapped by a fully-connected layer with a ReLU activation to a $7\times7$ tensor
with $64$ channels. We then have two blocks each consisting of a transposed
convolution, each halving the number of channels, followed by instance
normalization and a ReLU activation. Following the two blocks, there is a final
transposed convolution with a sigmoid activation. The discriminator mirrors this
structure: it uses three blocks each consisting of a convolution (to 64, 128,
and 256 channels respectively) followed by instance normalization and a ReLU
activation function. Finally a fully-connected layer maps the resulting tensor
to a scalar. See the code for full details. We train the GAN with a batch size
of 256; a learning rate of $10^{-3}$ for both the generator and the
discriminator; and the Adam optimizer with hyperparameters $\beta_1 = 0.5$ and
$\beta_2 = 0.999$. Each experiment is performed on a single NVIDIA V100 GPU (32G
HBM2 memory) and Intel Silver 4216 Cascade Lake @ 2.1GHz CPUs.

\subsection{CIFAR10}

We use the best-performing architecture reported by \citet{gong2019autogan}. All
hyperparameters are the same except for $\nadv$, which we vary. We use the
officially released code\footnote{https://github.com/VITA-Group/AutoGAN}. Each
GAN is trained on a single NVIDIA V100 GPU (32G HBM2 memory) and Intel Silver
4216 Cascade Lake @ 2.1GHz CPUs.

\subsection{CelebA}

We use a publicly-available implementation of
StyleGAN2\footnote{https://github.com/lucidrains/stylegan2-pytorch} with the
standard hyperparameter settings recommended by \citet{karras2020analyzing}.
Each experiment is performed on a single NVIDIA P100 GPU (12G HBM2 memory) and
Intel E5-2650 v4 Broadwell @ 2.2GHz CPUs.

\end{document}